\documentclass[a4paper,11pt]{article}

\usepackage{graphicx}  
\usepackage{url}       
\usepackage{times}
\usepackage{natbib}
\usepackage[margin=25mm]{geometry}

\usepackage{helvet}
\usepackage{courier}
\usepackage{tabu}
\usepackage{soul,color}
\usepackage{hyperref}
\usepackage{bussproofs}

\title{Towards Semantic Modeling of Contradictions and Disagreements: A Case Study of Medical Guidelines}
\date{}

\author{Wlodek Zadrozny\\
       Department of Computer Science, UNC Charlotte\\
       \texttt{wzadrozn@uncc.edu}
  \and Hossein Hematialam\\
       Department of Computer Science, UNC Charlotte\\
       \texttt{hhematia@uncc.edu}
  \and Luciana Garbayo\\
       Departments of Philosophy and Medical Education, U. of Central Florida\\
       \texttt{Luciana.Garbayo@ucf.edu}
}
\date{}

\begin{document}
\maketitle

\thispagestyle{empty}
\pagestyle{empty}

\begin{abstract}
We introduce a formal distinction between contradictions and disagreements in natural language texts, motivated by the need to formally reason about contradictory medical guidelines. This is a novel and potentially very useful distinction, and hasn't been discussed so far in NLP and logic. 
We also describe a NLP system capable of automated finding contradictory medical guidelines; the system uses a combination of text analysis and information retrieval modules. We also report positive evaluation results on a small corpus of contradictory medical recommendations.
\end{abstract}

\section{Introduction} 
This is a programmatic paper (and work in progress) motivated by the challenge of automatically identifying and representing contradictions in medical guidelines. In this paper we take the perspective of building a natural language understanding system that can properly represent disagreements. On the practical side, we expect this research to eventually result in a larger solution that can provide decision support for patients and physicians, and help identify and reason with contradictory advice in their specific cases. However, proposed solution apply more generally to natural language semantics.\\ 


\noindent\textbf{Contributions:} This paper makes the following novel contributions:
\begin{itemize}
\itemsep0em 
\item A novel formal analysis of types of contradictions in text. Namely, we introduce and formally characterize the distinction between \emph{contradictions} and \emph{disagreements}. This distinction is generally applicable to all semantic processing of natural language text, and is orthogonal to other typologies of contradictions, e.g. \cite{de2008finding}. 
\item A proposal for an architecture and a method for identifying contradictions and disagreement in medical guidelines (viewed as self-standing documents, and ignoring, in this paper, all epistemological issues in expert opinions) 
\item Preliminary results from an implemented system showing the feasibility of the proposed approach. 
\end{itemize}

\noindent\textbf{Motivation:}
Disagreements in medical guidelines raise uncertainty in disease screening and treatment. Uncertainty derived from the lack of guidelines consistency among different expert groups is confusing for patients, and also contributes to overdiagnosis . For example, the ACOG recommends that women over age 40 get a mammography annually, 
but the USPSTF recommends clinicians base screening decisions for women aged 40 to 49 on the women’s individual risk profile and preferences.  
We can see two different actions recommended by two guideline sets for women aged 40 to 49. We'll see other examples below.\\

\noindent\textbf{A Brief Overview of the Proposed Approach:} 
Clearly, dealing with multiple guidelines for the same condition can be reduced to analyzing the guidelines pairwise. Thus with two texts of such guidelines we propose the following:  
\begin{enumerate}
\itemsep0em
\item Identify candidate sentences pertaining to the same condition or the same action;
\item Compute candidate contradictions and disagreements (using techniques of information retrieval and statistical language modeling);
\item Identify the specific contradictions and disagreements computationally, using different, deeper modes of analysis based on semantic representation informed by \emph{ formal representations of disagreements and contradictions};
\item A method for automated 
automated reasoning with disagreements and contradictions in computational settings, focused on the identification areas of agreement and disagreement, including their provenances.
\end{enumerate}

\section{Formal representation of disagreement and contradiction}

We need a formal representation of contradictory guidelines in order to be able to reason about them. This section proposes, to our knowledge for the first time in formal semantics, a way to reason with partially contradictory information based on a formal distinction between disagreements and contradictions, and formalized using a combination of propositional calculus and lattice theory. 

Let's consider a few examples of actual sentences containing disagreements or contradictions. For clarity of exposition we will always present contradictions between pairs of documents, such as guidelines. 

\noindent\textbf{Example 1}. We will use an example from a CDC table comparing "Breast Cancer Screening Guidelines for Women" provided by seven different accredited medical bodies. 
\footnote{\url{ https://www.cdc.gov/cancer/breast/pdf/BreastCancerScreeningGuidelines.pdf} }
There we find contradictory recommendations for \c{\emph{"women aged 50 to 74 with average risk"}} coming from two (of the seven) different organizations):
\begin{quote}
(a) \emph{Screening with mammography and clinical breast exam annually.} \\
(b) \emph{Biennial screening mammography is recommended.}
\end{quote}

\noindent\textbf{Example 2.} Consider the question about the recommended number of minutes of physical activity. Again, the guidelines might differ
$\colon$ One organization recommending a minimum of 150 minutes per week, and another 150-300 minutes week (we simplify the recommendation a bit here).
Clearly, someone exercising 30 min per day, 6 days a week satisfies both guidelines. The guidelines don't agree 100\%, but intuitively they are not 100\% contradictory either. \\

\noindent\textbf{Disagreements vs contradictions: }
To capture the intuitive distinction between Examples 1 and 2, we say that two guidelines are \emph{contradictory} if it is impossible for bothquidelines to be  followed. And two guidelines are in \emph{disagreement} if there are patients where the two guidelines are possible to followed, and patients for which this is impossible. 
As it turns out we can represent this distinction formally, in logic, making it broadly applicable in semantics, using the following idea:
\begin{itemize}
\itemsep0em
\item	\emph{Contradiction} is present if there is no model for the joint theory expressed in two text segments (coming from different guidelines)
\item	\emph{Disagreement} is present, if the sets of models, for the predicates present in both text segments, are different for each segment, but a model can be created satisfying both segments.\\
\end{itemize}

\noindent\textbf{Formalization: }
We start by assuming that at least initially we do not need full power of first order logic (FOL) or a stronger logical system. So the basis of our representation will be a formal language of propositions. Thus we do not have variables or quantifiers. However, to be able to reflect the disagreements we need to augment it with a representation of parameters. For example, we would like to be able distinguish between a recommendation of a minimum of  30 min of daily exercise and another one of 20 minutes; and at the same time we need to be able to notice that both recommendations pertain to the recommended dose of exercise.

To this end we assume that our representation language contains propositional symbols $p,q,r, ..., p_1,$ $ p_2, ...$ and symbols representing parameters 
$a, b, c, a_1, a_2,$ etc.   
We have special parameters $o_1, o_2, ...$ which later will be used to represent the provenance of recommendations. This will allow us to find the sources of contradictions and disagreements.

We assume the standard inference rules of propositional logic. (The added parameters obviously don't extend the power of the system beyond propositional calculus). 
To reason about disagreements we will need to introduce additional rules of inference. \\

\noindent\textbf{Example 1 continued: } 
Let $p$ stand for \emph{screening mammography is recommended}; $o_1, o_2$ represent the provenance of the recommendations (a) and (b) respectively; and $a, b$ stand for \emph{annually} and \emph{biennally}. We then have the formal representation of the respective guidelines as  $p(a, o_1)$  and $p(b, o_2)$ .

We need means to formally represent the fact that these guidelines are contradictory. This cannot simply be due to having different constants/parameters inside the parentheses (and ignoring the $o$'s). To see that consider a similar representation of doses of daily recommended exercise. Here, we would also have two distinct provenances and two distinct values, however, intuitively we could recognize a disagreement and not a contradiction, since anyone exercising 30 min or more is also exercising 20 min or more.

To proceed we need to make two additional assumptions, namely that no particular guidelines document can have internal contradictions. That is, the set of all  
$p_i(a^i_j, o) $
for a particular $o$ is never contradictory (viewed as statements in classical propositional logic). 

And the second assumptions is that the parameters come in different sorts, which we will represent by capital letters followed by a colon, e.g. $A:a_1$.  More importantly, we assume elements of any particular sort form a \emph{lattice} (or at least\emph{ meet semi-lattice}). That is, for any set of parameters of a particular sort (e.g. time, duration, dosage, etc.) $a_1 \wedge a_2$ is defined, and every such lattice has a minimal element $\bot$. 

In the example representations of mammography the meet of \emph{biennial} and \emph{annual} is $\bot$. However, the meet of "20 min or more" and "30 min or more" is the latter.  This mechanism allows us to make a formal distinction between contradictions and disagreements.

$p(A\colon a_1, o_1) $ and   $p(A\colon a_2, o_2) $      are         \emph{contradictory} if $a_1 \wedge a_2  = \bot$ 

$p(A\colon a_1, o_1) $ and   $p(A\colon  a_2, o_2) $         \emph{disagree} if taking  $a_1 \wedge a_2  = a$, we have $a \ne \bot$ and either 
$a \ne a_1$ or $a \ne a_2$. \\

These definitions are naturally extended to multi-parameter case by defining the contradiction as a situation, where the meet of at least one type of parameters is $\bot$; and the disagreement, when there's no contradiction, and at least one type of parameter contains a disagreement. 

To do some elementary reasoning about disagreement we need an inference rule capable of relating two formulas with different parameters, and to keep track of provenances we need to allow propositions with multiple labels, e.g.  $\{o_1, o_2\}$. 
This is nicely combined in a single inference rule: 



\begin{prooftree}
\AxiomC{$p(A\colon a_1, o_1) , p(A\colon a_2, o_2) $}
\RightLabel{Lattice $\wedge$ }
\UnaryInfC{$p(A\colon a_1 \wedge a_2, \{o_1,o_2\})$}
\end{prooftree}
If the formula $p$ has more than one parameter, we apply this rule for each parameter separately. \\ 
\noindent With this inference rule we are getting the following:
\begin{itemize}
\itemsep0em 
\item The set of derivable (using the above rule) contradictory propositions corresponds to the ones that have $\bot$ as at least one parameter.
\item The set of disagreements corresponds to the derivable propositions with two or more provenance parameters and a disagreement for one or more sorts.
\item For any fixed provenance $o$ we have the full power of power of inference rules of propositional calculus applied to sentences of the form $p(a,o)$ where $a$ stands for a collection of parameters of different sorts. 
\end{itemize}

\section{Finding contradictions and disagreements}

Having solved the problem of formally representing contradictions and disagreement, and having created a formal method of keeping track of their provenances, we now focus on the language understanding part. 

The results presented in this section are preliminary in two ways: First, we have not completed a translation from a semantic representation produced by NLP tools to a logical form amenable to reasoning with the parameterized propositional logic of the previous section. We assume this can be done using existing methods, as described in publications ranging from standard textbooks (\cite{nltkbook}) to complex NLP architectures (\cite{mccord2012deep}). Obviously we intend to prove that this indeed is possible.  
Second, our methods for finding contradictions and disagreements, even though not trivial, very likely can be improved. Nevertheless the results are promising. 

\begin{figure}
  \includegraphics[scale=0.47]{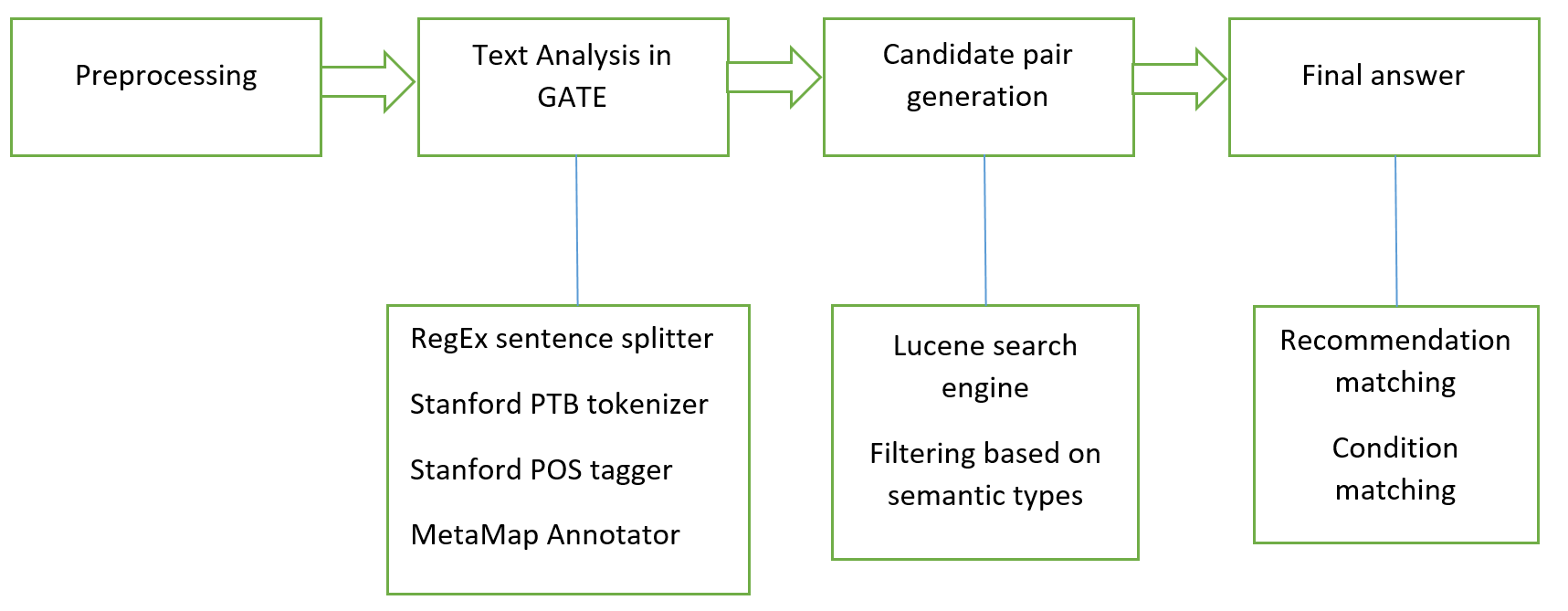}
  \caption{The architecture used to evaluate extraction of contradictions in medical guidelines}
  \label{fig:TextExtraction}
\end{figure}

Figure \ref{fig:TextExtraction} shows a novel architecture consisting of several known well components. We follow the approach presented in the Introduction:  we use the text analysis tools for feature generation and concept identification. The Lucene search engine is used to for finding similar sentences based on indexed semantic features (and words). For examples, given the query "mammography is recommended for women age 40-49" we search multiple guidelines and identify sentences for further analysis. This analysis done through recommendation matching ('mammography recommended') and condition matching ('age 40-49', or 'age') allows the system to decide if given guidelines document recommends a procedure or not. Similarly, we can identify partial matches, e.g. 'age 40-49' and 'age over 40'. 

\noindent\textbf{Evaluation:} At this point we only evaluated this method on finding agreement and disagreement on twelve example recommendations sentences and breast cancer screening guidelines produced by seven different medical organizations. 
\footnote{\url{ https://www.cdc.gov/cancer/breast/pdf/BreastCancerScreeningGuidelines.pdf} } This gives us only 84 data points. However, the results are promising: The system produced only four errors (two false positives and two false negatives), thus on this -- admittedly, simple -- data set achieved an impressive accuracy of 95\%. 

\section{Summary, comparisons, and ongoing work}
\textbf{Summary: } Motivated by analysis of medical guidelines we introduced the formal distinction between disagreements and contradictions. We presented a new system for finding both, and results of a preliminary evaluation. The new formal representation and the general architecture of the system are potentially broadly applicable to NLP, for example to question answering, where an answer can be extracted from texts that disagree on details, but broadly provide the same answer or recommend the same action. 
Many obstacles remain: we do not have reliable ways of converting longer texts intended for human reading into semi-structured representation suitable for text mining (for example dealing with tables); in evaluation we used simple sentences, but texts might contain information in multiple sentences, and thus increasing the difficulty of matching, and necessitating the need to combine partial information. 
And while solutions to these problems exist, they are not perfect, and will likely decrease the accuracy of the system.

\noindent\textbf{Comparisons: } We want to acknowledge prior work on representing contradictions in NLP, e.g. \cite{de2008finding} and \cite{kloet2013twoShort}. The former containing a taxonomy of linguistic expressions of contradiction, potentially useful when dealing with the linguistic diversity of the guidelines. The latter showing methods for large scale acquisition of contradictory patterns. We believe such distributional methods might add coverage to our approach and complement the IR method we are currently using.  Neither of these works makes a formal distinction between contradictions and disagreements.  On the formal side, 
clearly there is a big body of work on contextualizing the truth of propositions, for example in modal logic and para-consistent logics. \footnote{\url{https://plato.stanford.edu/entries/possible-worlds/}{, and} \url{https://plato.stanford.edu/entries/logic-paraconsistent/}}. We even found, accidentally, a  1969 paper on 'topological logic' \cite{rescher1969topological}, which parameterizes propositions in several ways, but does not use lattices of parameters -- this seems to be our original contribution. 

\noindent\textbf{Ongoing work: } In addition to the NLP work, we are exploring opportunities to apply the results of disagreement analysis in medical practice. On the language processing side, we will be evaluating the current system on linguistically more complex sets of guidelines. This will require additional text prepossessing and dealing with information spread out in multiple sentences. We have done before work \cite{hematial2016s} on extracting condition and action expressions in medical guidelines for sinusitis, hypertension and asthma, with acceptable results. We plan to evaluate our current work by adding guidelines for the same conditions issued by other medical organizations.

\textbf{Acknowledgements}. The authors would like to thank James Stahl of the Department of Internal Medicine, Dartmouth College, for discussions of contradictions in medical guidelines. 
\noindent
\bibliographystyle{chicago}
\bibliography{coguidelines}

\end{document}